# SSGAN: Secure Steganography Based on Generative Adversarial Networks

Haichao Shi[a,b], Jing Dong[c], Wei Wang[c], Yinlong Qian[c], Xiaoyu Zhang[a*]

[a]Institute of Information Engineering, Chinese Academy of Sciences
[b] School of Cyber Security, University of Chinese Academy of Sciences
[c]Center for Research on Intelligent Perception and Computing, National Laboratory of Pattern Recognition, Institute of Automation, Chinese Academy of Sciences

**Abstract.** In this paper, a novel strategy of Secure Steganograpy based on Generative Adversarial Networks is proposed to generate suitable and secure covers for steganography. The proposed architecture has one generative network, and two discriminative networks. The generative network mainly evaluates the visual quality of the generated images for steganography, and the discriminative networks are utilized to assess their suitableness for information hiding. Different from the existing work which adopts Deep Convolutional Generative Adversarial Networks, we utilize another form of generative adversarial networks. By using this new form of generative adversarial networks, significant improvements are made on the convergence speed, the training stability and the image quality. Furthermore, a sophisticated steganalysis network is reconstructed for the discriminative network, and the network can better evaluate the performance of the generated images. Numerous experiments are conducted on the publicly available datasets to demonstrate the effectiveness and robustness of the proposed method.

## 1 Introduction

Steganography is the task of concealing a message within a medium such that the presence of the hidden message cannot be detected. It is one of the hot topics in information security and has drawn lots of attention in recent years. Steganography is often used in secret communications. Especially in the fast-growing social networks, there are an abundance of images and videos, which provide more opportunities and challenges for steganography. Therefore, the design of a secure steganography scheme is of critical importance.

How to design a secure steganography method is the problem that researchers have always been concerned about. Existing steganographic schemes usually require the prior of probability distribution on cover objects which is difficult to obtain in practice. Conventionally, the steganography method is designed in a heuristic way which does not take the steganalysis into account fully and automatically. For the sake of the steganography safety, we consider the steganalysis into the design of steganography.

* Corresponding author: Xiaoyu Zhang, zhangxiaoyu@iie.ac.cn

At present, the image-based steganography algorithm is mainly divided into two categories. The one is based on the spatial domain, the other is based on the DCT domain. In our work, we focus on the spatial domain steganography.

Least Significant Bit (LSB) [11] is one of the most popular embedding methods in spatial domain steganography. If LSB is adopted as the steganography method, the statistical features of the image are destroyed. And it is easy to detect by the steganalyzer. For convenience and simple implementation, the LSB algorithm hides the secret to the least significant bits in the given image's channel of each pixel. Mostly, the modification of the LSB algorithm is called ±1-embedding [2]. It randomly adds or subtracts 1 from the channel pixel, so the last bits would match the ones needed. So we consider the ±1-embedding algorithm in this paper.

Besides the LSB algorithm, some sophisticated steganographic schemes use a distortion function which is used for selecting the embedding localization of the image. We called them the image content-adaptive steganography. These algorithms are the most popular and the most secure image steganography in spatial domain, such as HUGO (Highly Undetectable steGO), WOW (Wavelet Obtained Weights), S-UNIWARD, etc..

HUGO [12] is a steganographic scheme that defines a distortion function domain by assigning costs to pixels based on the effect of embedding some information within a pixel. It uses a weighted norm function to represent the feature space. HUGO is considered to be one of the most secure steganographic techniques, which we will use in this paper to demonstrate our method's security. WOW (Wavelet Obtained Weights) [8] is another content-adaptive steganographic method that embeds information into a cover image according to textural complexity of regions. In WOW shows that the more complex the image region is, the more pixel values will be modified in this region. S-UNIWARD [9] introduces a universal distortion function that is independent of the embedded domain. Despite the diverse implementation details, the ultimate goals are identical, i.e. they are all devoted to minimize this distortion function, to embed the information into the noise area or complex texture, and to avoid the smooth image coverage area.

So far as we know, when people design steganography algorithm, they usually heuristically consider the steganalysis side. For example, the message should embed into the noise and texture region of image which is more secure. In this paper, we propose a novel SSGAN algorithm, which implements secure steganography based on the generative adversarial networks. We consider it under adversarial learning framework, inspired by the work of Denis and Burnaev [5]. We use WGAN [3] to improve the security of steganography by generating more suitable covers. In the proposed SSGAN, covers are generated firstly using the generative network. Then we adapt the state-of-the-art embedding algorithm, like HUGO, to embed message into the generated image. Finally, we use the GNCNN [17] to detect on images whether there is a steganographic operation.

The contributions of our work can be concluded as follows:

**Perceptibility.** In this paper, we use WGAN instead of DCGAN to generate cover images to achieve generative images with higher visual quality and ensure faster training process.

**Security.** We use a more sophisticated network called GNCNN to assess the suitableness of the generated images instead of the steganalysis network proposed by [5].

**Diversity.** We also use GNCNN to compete against the generative network, which can make the generated images more suitable for embedding.

The rest of the paper is structured as follows: In Section 2, we discuss the related work of adversarial learning and elaborate the proposed method. In Section 3, experiments are conducted to demonstrate the effectiveness and security of the proposed method. In Section 5, we draw conclusions.

## 2     Secure Steganography Based on Generative Adversarial Networks

### 2.1    Adversarial Learning

Adversarial learning using game theory, and is combined with unsupervised way to jointly train the model. The independent model is trained to compete with each other, iteratively improving the output of each model. In Generative Adversarial Networks, the generative model G tries to train the noise to samples, while the discriminative model D tries to distinguish between the samples output by G and the real samples. Based on fooling the D, the weight of G is updated, and at the same time, the D's weight is updated by distinguishing between the fake and real samples.

Recent years, GANs have been successfully applied to image generation tasks [14] using convolutional neural networks (CNNs) for both G and D. But in traditional GANs, they are considered difficult to train. Because there is no obvious relationship between the convergence of the loss function and the sample quality. Typically, people choose to stop the training by visually checking the generated samples. So, a design arises recently, WGAN [3], using the Wasserstein distance instead of the Jensen-Shannon divergence, to make the data set distribution compared with the learning distribution from G. Obviously, they show that the sample quality is closely related with the network's convergence and the training rate is really improved.

Adversarial training has also been applied to steganography. The adversarial training process can be described as a minimax game:

$$\min_G \max_D J(D,G) = E_{x \sim p_{data}(x)} \log(D(x)) + E_{z \sim p_{noise}(z)} \log(1 - D(G(z))) \quad (1)$$

where *D(x)* represents the probability that x is a real image rather than synthetic, and G(z) is a synthetic image for input noise z. In this process, there are two networks, the G and the D, trained simultaneously:

- Generative Network, its input is a noise z from the prior distribution $p_{noise}(z)$, and transform it from the data distribution $p_{data}(x)$, to generate a data sample which is similar to $p_{data}(x)$.

- Discriminative Network, its input are the real data and the fake data generated from the Generative Network, and determine the difference between the real and fake data samples.

To solve the minimax problem, in each iteration of the mini-batch stochastic gradient optimization, we first perform the gradient ascent step on D and then perform the gradient descent step on G. So we let $\omega_N$ represents the neural network N, then we can see the optimization step:

- We let the D fixed to update the model G by $\omega_G \leftarrow \omega_G - \gamma_G \nabla_G J$

where

$$\nabla_G J = \frac{\partial}{\partial \omega_G} E_{z \sim p_{noise}(z)} \log(1 - D(G(z, \omega_G), \omega_D)) \tag{2}$$

- We let the G fixed to update the model D by $\omega_D \leftarrow \omega_D + \gamma_D \nabla_D J$

where

$$\nabla_D J = \frac{\partial}{\partial \omega_D} \{E_{x \sim p_{data}(x)} \log(D(x, \omega_D)) + E_{z \sim p_{noise}(z)} \log(1 - D(G(z, \omega_G), \omega_D))\} \tag{3}$$

In this paper, we use WGANs to verify the advantages of generating and discriminating the image using adversarial training process.

## 2.2 Model Design

We introduce a model that we called SSGAN, which contains a generative network, and two discriminative networks. The model can be described as Fig. 1:

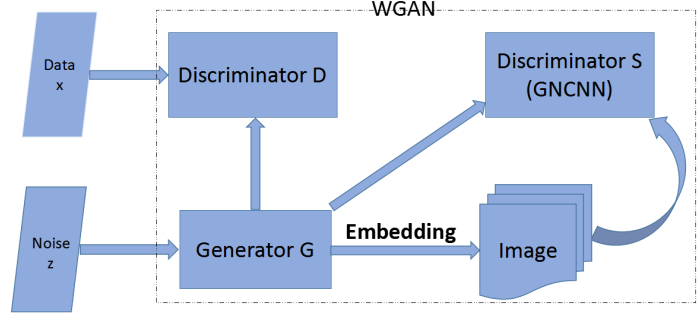

**Fig. 1.** The SSGAN model

Since we want the G to generate realistic images that could be used as secure covers for steganography, we force G to compete against the D and S at the same time. We use $S(x)$ to represent the output of the steganalysis network, then the game can be shown as follows:

$$\min_G \max_D \max_S J = \alpha \left( E_{x \sim p_{data}(x)} \log(D(x)) + E_{z \sim p_{noise}(z)} \log\left(1 - D(G(z))\right) \right)$$

$$+ (1-\alpha) E_{z \sim p_{noise}(z)} \left[ \log S\left(Stego(G(z))\right) + \log\left(1 - S(G(z))\right) \right]$$

$$\tag{4}$$

To control the trade-off between the realistic of the generated images and the evaluation of the steganalysis, we use a convex combination which includes the D and S network with parameters α ∈ [0，1]. And we show that when we give α ≤ 0.7, the results are closer to the noise.

### 2.2.1 Generator G

For the generator G, it is used to generate the secure covers. And we use a fully connected layer, and four fractionally-strided convolution layers, and then a Hyperbolic tangent function layer. This network structure can be described as Fig. 2:

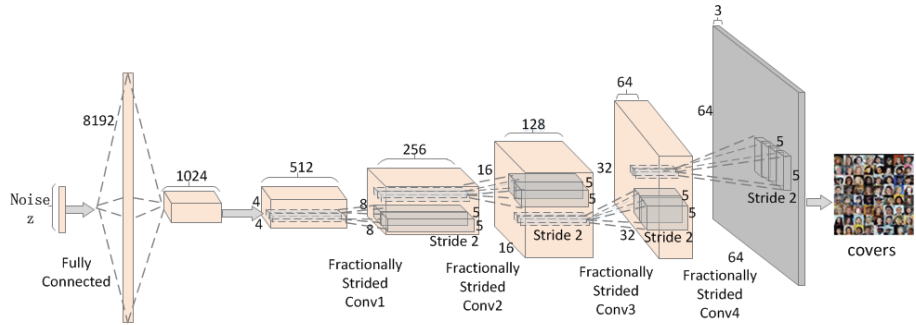

**Fig. 2.** The generative network structure

### 2.2.2 Discriminator D

For the discriminator D, it is used to evaluate the visual quality of the generated images. And we use four convolutional layers, and then a fully connected layer. This network structure can be described as Fig. 3:

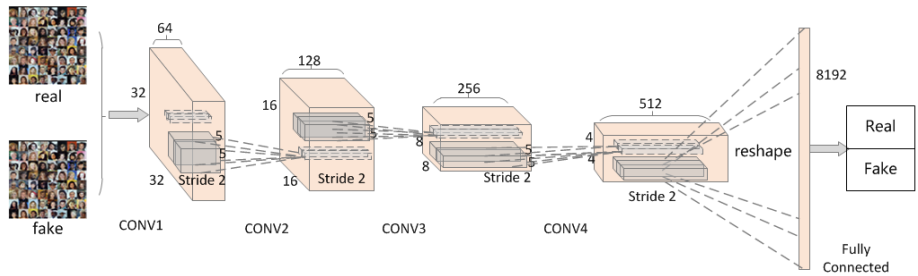

**Fig. 3.** The discriminative network structure

### 2.2.3 Discriminator S

For the S network, it is used to assess the suitableness of the generated images. And we first use a predefined high-pass filter to make a filtering operation, which is mainly for steganalysis. And then four convolutional layers. Finally we use a classification layer, which includes several fully connected layers. This network structure can be described as Fig. 4.

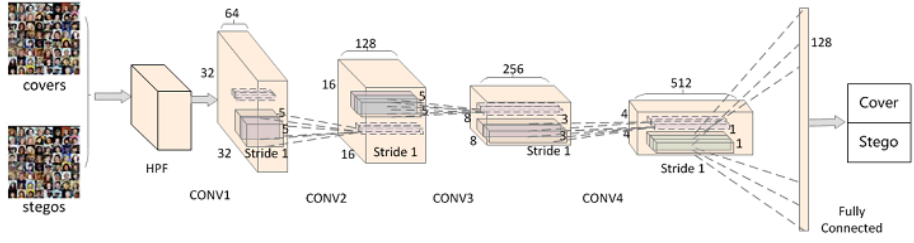

**Fig. 4.** The steganalysis network structure

### 2.2.4 Update Rules

And we can show that the SGD update rules:

- For the generator G: $\omega_G \leftarrow \omega_G - \gamma_G \nabla_G J$ it is calculated by

$$\nabla_G J = \frac{\partial}{\partial \omega_G} E_{z \sim p_{noise}(z)}\left[\log(1 - D(G(z, \omega_G), \omega_D))\right] +$$

$$\frac{\partial}{\partial \omega_G}(1-\alpha) E_{z \sim p_{noise}(z)}\left[\log\left(S(Stego(G(z, \omega_G), \omega_S))\right)\right] +$$

$$\frac{\partial}{\partial \omega_G}(1-\alpha) E_{z \sim p_{noise}(z)}\left[\log(1 - S(G(z, \omega_G), \omega_S))\right] \quad (5)$$

- For the discriminator D: $\omega_D \leftarrow \omega_D + \gamma_D \nabla_G J$ it is calculated by

$$\nabla_G J = \frac{\partial}{\partial \omega_D}\{E_{z \sim p_{data}(x)}[\log D(x, \omega_D)] + E_{z \sim p_{noise}(z)}[\log(1 - D(G(z, \omega_G), \omega_D))]\} \quad (6)$$

- For the discriminator S: $\omega_S \leftarrow \omega_S + \gamma_S \nabla_S J$ it is calculated by

$$\nabla_S J = \frac{\partial}{\partial \omega_S} E_{z \sim p_{noise}(z)}\left[\log S(Stego(G(z, \omega_G)), \omega_S) + \log(1 - S(G(z, \omega_G), \omega_S))\right] \quad (7)$$

We update G to not only maximize the errors of D, but the normalization errors of D and S.

## 3 Experiments

### 3.1 Data preparation

All experiments are performed in TensorFlow [1], on a workstation with a Titan X GPU. In our experiments, we use the CelebA dataset which contains more than 200,000 images.

We pre-process the image, and all images are cropped to $64 \times 64$ pixels.

For the purpose of the steganalysis, we use 90% of the data to construct a training set, and regard the rest as testing set. The training set is denoted by TRAIN, and the testing set is denoted by TEST. We use Stego(x) to represent the steganographic algorithm used to hide information. Two datasets are involved in the experiments. One is TRAIN + Stego(TRAIN), where Stego(TRAIN) is the training set embedded in some secret information, and the other is TEST + Stego(TEST). Finally, we got 380,000 images for steganography training, 20,000 for testing. In order to train the model, we use 200,000 cropped images to generate images. After seven epochs, the images generated by our generative model and the model proposed in [5], denoted by SGAN, are shown in the Fig. 5.

Meanwhile, we use the LSB Matching algorithm which is ±1-embedding algorithm with a payload size to 0.4 bits per pixel to embed information, which we use a text from casual articles.

The experimental results are as follows, the visual quality of images generated by our SSGAN model are higher than its counterpart.

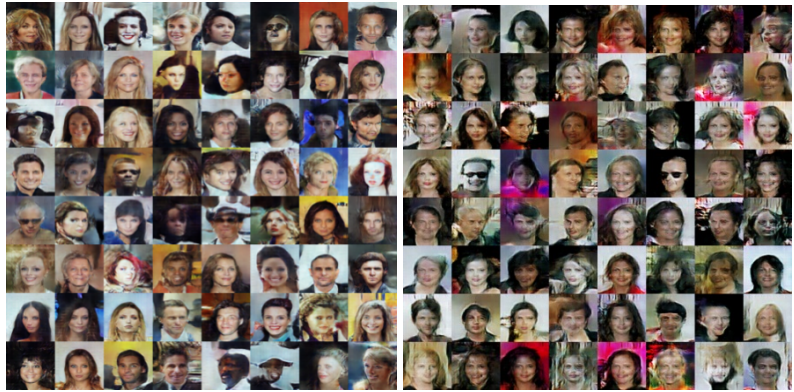

**Fig. 5.** Examples of images, generated by SSGAN and SGAN after training for 7 epochs on the CelebA dataset, the left is generated by SSGAN, the right is generated by SGAN.

Experimental results show that through the use of WGAN, the convergence speed is faster than DCGAN, and the effect is more obvious, as is shown in Table 1.

**Table 1.** The contrast of two methods' time for running for seven epochs

| Method | Time(mins) |
|--------|------------|
| SSGAN  | 227.5      |
| SGAN   | 240.3      |

### 3.2 Experimental Setup

The generative network G is designed with one fully connected layer, four Fractionally − Strided $Conv2D \rightarrow Batch\ Normalization \rightarrow Leaky\ ReLU$ layers, and one Hyperbolic tangent function layer. The discriminative network D comprises four $Conv2D \rightarrow Batch\ Normalization \rightarrow Leaky\ ReLU$ layers, one fully connected layer. As for the steganalyser network S, we use a predefined high-pass filter to make a filtering operation, which is kept fixed while training, and four convolutional layers corresponding to three kinds of operations, i.e. convolution, non-linearity, and pooling, followed by a fully connected layer and a softmax layer for classification.

We train the model using RMSProp optimization algorithm with the learning rate $2 \times 10^{-4}$ and update parameters $\beta_1 = 0.5$ and $\beta_2 = 0.99$. We update the weights of D and S once, while update weights of G twice in each mini-batch. In steganalysis, we call S the Steganalyser. In addition to the S network, we also use an independent network which is called S* [17]. To make the network S effective, we also use the filter in [17], which defines as follows:

$$F^{(0)} = \begin{pmatrix} -1 & 2 & -2 & 2 & -1 \\ 2 & -6 & 8 & -6 & 2 \\ -2 & 8 & -12 & 8 & -2 \\ 2 & -6 & 8 & -6 & 2 \\ -1 & 2 & -2 & 2 & -1 \end{pmatrix}$$

We also train the S* network using RMSProp optimization algorithm with the learning rate $5 \times 10^{-6}$ and update parameters $\beta_1 = 0.9$ and $\beta_2 = 0.99$. And we use the format of binary cross-entropy to calculate the loss.

We conduct experiments on real images and generative images respectively, and the results are as follows:

In our first experiment, we compare the security of the generated images with that of real images. We embed secret message into real images and use the steganalysis network on real images first, then we embed secret message into generative images and use the steganalysis network on generative images.

We train the S network on real images and from the results we conclude that even if the usual WGAN can generate a synthetic image, it can be easily detected by a steganalyser. And our method increase the results of the classification error rates, which means the images generated by our model is more secure to be the carrier of the steganography images.

**Table 2.** Accuracy of the steganalyser trained on real images

| Type of Images | SSGANs | SGANs |
|---|---|---|
| Real images | 0.87 | 0.92 |
| Generated images | 0.72 | 0.90 |

We also use HUGO steganography scheme on real images and generative images.

In our second experiment, we investigate the security of generated images under different seed values. We conduct the experiment on generative images generated by different setups.

In this group of experiments, we use Qian's network [17] which is called steganalyser[1] S* on generated images generated by SSGAN. The input is the prior noise distribution $p_{noise}(z)$ for some fixed seed value. We test the S* on images. The experimental setups are as follows:

S1. We use same seed value;

---

[1] Note that for the issue of image size, we simply modify the structure of GNCNN as SSGAN's steganalyser.

S2. We use some randomly selected seed value.

S3. We use the same seed value, as in S2, and we additionally tune the WGAN model for several epochs.

Table 3. Trained on generated images according to experimental conditions S1-S3

| Experimental Conditions | Accurcy |
|---|---|
| S1 | 0.87 |
| S2 | 0.72 |
| S3 | 0.71 |

As is shown in Table 3, we can see that through using different seed values when generating images, can make it easier to deceive the steganalysis network.

### 3.3 Discussion

As demonstrated in the experiments, the method proposed in [5] has some limitations, the experiments show that the steganography is not secure enough. The experiments show that the steganalysis network in Denis's article is suitable for embedding with the same key, but when using the random key, their network is not so useful. So we use GNCNN [17]. And we have confirmed that the steganalysis network they used was not so useful. Our results show that on the one hand, the generated images are more difficult to detect, indicating that the security performance is higher. On the other hand, the generated images' visual quality is better and more realistic.

## 4 Conclusion And Future Work

In this paper, we introduce generative adversarial networks for steganography to generate more suitable and secure covers for steganography. Based on the WGANs, we have proposed a model called SSGAN for steganography. The proposed model is efficient to generate images, which have higher visual quality. And our model is suitable for embedding with the random key. Mostly, it can generate more secure covers for steganography. We have evaluated the performance of our model using CelebA datasets. Results show the effectiveness of the SSGAN model through the classification accuracy, and we think it could be used for adaptive steganographic algorithm for social network in the future. We believe that, by exploring more steganography properties, better performance can be achieved.

## 5 Acknowledgement

This work is supported by the National Natural Science Foundation of China (No.61501457, U1536120, U1636201, 61502496) and the National Key Research and Development Program of China (No.2016YFB1001003).